\documentclass[11pt,a4paper]{article}

\usepackage{authblk}
\usepackage[hyperref]{acl2019}
\usepackage{times}
\usepackage{latexsym}
\usepackage{gb4e}
\noautomath
\usepackage{epigraph}
\usepackage{url}
\usepackage{tabularx}
\usepackage{caption}
\usepackage{subcaption}
\usepackage{graphicx}

\aclfinalcopy % Uncomment this line for the final submission
\begin{document}

%\title{Countering Hate: Nichesourcing a large-scale multilingual dataset of responses to Online Hate content (tentative)}
\title{CONAN - COunter NArratives through Nichesourcing: \\ a Multilingual Dataset of Responses to Fight Online Hate Speech }
%\title{CONAN:  a multilingual dataset of COunter NArratives Nichesourced to experts to fight Hate Speech (tentative)}

% \author{First Author \\
%  Affiliation / Address line 1 \\
%  Affiliation / Address line 2 \\
%  Affiliation / Address line 3 \\
%  \texttt{email@domain} \\\And
%  Second Author \\
%  Affiliation / Address line 1 \\
%  Affiliation / Address line 2 \\
%  Affiliation / Address line 3 \\
%  \texttt{email@domain} \\}
  
%\author{Yi-Ling Chung\inst{1,2}\\
%  %FBK, Trento, Italy / Address line 1 \\
%  \texttt{ychung@fbk.eu} \\\And
%  Elizaveta Kuzmenko\institute{2} \\
%  %Affiliation / Address line 1 \\
%  \texttt{email@domain} \\\And
%  Serra Sinem Tekiroglu\institute{1} \\
%  \texttt{tekiroglu@fbk.eu} \\\And
%  Marco Guerini\institute{1} \\
%  \texttt{guerini@fbk.eu} \\
%  }

%\author{ Yi-Ling Chung, Elizaveta Kuzmenko Serra Sinem Tekiro\u{g}lu  Marco Guerini \\
% Fondazione Bruno Kessler, Via Sommarive 18, Povo, Trento, Italy\\
%  \texttt{ychung@fbk.eu,email@domain,tekiroglu@fbk.eu, guerini@fbk.eu}
% }

\renewcommand\Authfont{\bfseries}

\author[1,2]{Yi-Ling Chung}
\author[2]{Elizaveta Kuzmenko}
\author[1]{Serra Sinem Tekiro\u{g}lu}
\author[1]{Marco Guerini}

\affil[1]{Fondazione Bruno Kessler, Via Sommarive 18, Povo, Trento, Italy
\protect\\ \texttt{ychung@fbk.eu, tekiroglu@fbk.eu, guerini@fbk.eu}}
\affil[2]{University of Trento, Italy \protect\\
%Via Calepina, 14, Trento
\texttt{elizaveta.kuzmenko@studenti.unitn.it}}

%\author{Yi-Ling Chung\inst{1,2} \And
%  \texttt{ychung@fbk.eu} \\\And
%  Elizaveta Kuzmenko\inst{2} \And
  %\texttt{email@domain} \\\And
%  Serra Sinem Tekiroglu\inst{1}  \And
%  %\texttt{tekiroglu@fbk.eu} \\\And
%  Marco Guerini\inst{1} }
  %\texttt{guerini@fbk.eu} \\}

%\date{}

%\begin{document}
\maketitle
\begin{abstract}
Although %the proliferation  of  online  hate  speech has fostered 
there is %now \sst{why now?} 
an unprecedented effort to provide adequate responses in terms of laws and policies to hate content on social media platforms, %such content is continuously evolving and adapting, making 
dealing with hatred online is still a tough problem. %Apart from the fact that data collection is often hampered by privacy policies and data deletion, The strategy of automatically recognizing inappropriate content so to signal it, 
Tackling hate speech in the standard way of content deletion or user suspension may be charged with censorship and overblocking. %\sst{We are talking about data collection here, but in the next "the alternate strategy" is for opposing it. How opposing will help data collection? And if we use "automatically recognizing", it seems like our paper is on recognizing at the first glance. What do you think?}
One alternate strategy, that has received little attention so far by the research community, is to actually oppose hate content with counter-narratives (i.e. informed textual responses). In this paper, we describe the creation of the first large-scale, multilingual, expert-based dataset of hate speech/counter-narrative pairs. This dataset has been built with the effort of more than 100 operators from three different NGOs that applied their training and expertise to the task. % collected during several NGO's operators training sessions \sst{why are we talking about training sessions here??}.
Together with the collected data we also provide additional annotations about expert demographics, hate and response type, and data augmentation through translation and paraphrasing. Finally, we provide initial experiments to assess the quality of our data. 
\end{abstract}

\section{Introduction}
%%\epigraph{\textit{Darkness cannot drive out darkness; only light can do that. Hate cannot drive out hate; only love can do that.}} {Martin Luther King, Jr.}

Together with the rapid growth of social media platforms, the amount of user-generated content is steadily increasing. At the same time, abusive and offensive language can spread quickly and is difficult to monitor. Defining hate speech is challenging for the broadness and the nuances in cultures and languages. For instance, according to UNESCO hate speech refers to ``expressions that advocate incitement to harm based upon the target’s being identified with a certain social or demographic group" \cite{gagliardone2015countering}.
%In this regard, hate speech is commonly defined as a denigrating expression, targeting a specific person or community, that causes threats and crimes to society \cite{karst200encyclopedia}. %This paper releases a first ever multilingual dataset for counter-narrative to combat hate speech.

Victims of hate speech are usually targeted because of various aspects such as gender, race, religion, sexual orientation, physical appearance. For example, Sentence \ref{ex_a} shows explicit hostility towards a specific group with no reasons explained\footnote{It is crucial to note that this paper contains examples of language which may be offensive to some readers. They do not represent the views of the authors.}. %while some hate speech contains implicit but extreme hostility against hate targets. %\sst{Example?? -> The vast majority of muslims are unemployed and refuse to work. / Muslims have nothing to contribute to our society. }. %Consider the example of hate speech below: 
%\vspace{2mm}
\begin{exe}
\ex \label{ex_a} I hate Muslims. They should not exist.
\end{exe}
%\vspace{2mm}

Online hate speech can deepen prejudice and stereotypes \cite{citron2011intermediaries} and bystanders may receive false messages and consider them correct. Although Social Media Platforms (SMP) and governmental organizations %have recognized the increasingly common problem and 
have elicited unprecedented attention to take adequate actions against hate speech by implementing laws and policies  \cite{gagliardone2015countering}, %Even if countries and companies are implementing laws and policies against hate speech, 
they do not seem to achieve the desired effect, since hate content is continuously evolving and adapting, making its identification 
a tough problem \cite{davidson2017automated}.  %\textbf{CITE!!}
% one possible reason lie in the difficulties of distinguishing hate speech and offensive language \cite{davidson2017automated}.

The standard approach used on SMPs to prevent hate spreading is the suspension of user accounts or deletion of hate comments, while trying to weigh the right to freedom of speech. Another strategy, which has received little attention so far, is to use counter-narratives. A counter-narrative (sometimes called counter-comment or counter-speech) is a response that provides non-negative feedback %to withstand hate speech 
through fact-bound arguments and is considered as the most effective approach to withstand hate speech \cite{benesch2014countering, schieb2016governing}. In fact, it preserves the right to freedom of speech, counters stereotypes and misleading information with credible evidence. It can also alter the viewpoints of haters and bystanders, by encouraging the exchange of opinions and mutual understanding, and can help de-escalating the conversation. A counter-narrative such as the one in Sentence \ref{ex_b} is a non-negative, appropriate response to Sentence \ref{ex_a}, while the one in \ref{ex_c} is not, since it escalates the conversation.

\begin{exe}
\ex \label{ex_b} Muslims are human too. People can choose their own religion.
%\phantom{(b}(c) I guess you hate yourself, isn't it?
\ex \label{ex_c} You are truly one stupid backwards thinking idiot to believe negativity about Islam. %\mg{change a bit}
\end{exe}

In this respect, some NGOs are tackling hatred online by training operators to monitor SMPs and to produce appropriate counter-narratives when necessary. Still, manual intervention against hate speech is a toil of Sisyphus, and automatizing the countering procedure would increase %increment 
the efficacy and effectiveness of hate countering \cite{munger2017tweetment}.
 
As a first step in the above direction, %automatizing counter-narrative production
we have nichesourced the collection of a dataset of counter-narratives to 3 different NGOs. Nichesourcing is a specific form of outsourcing that harnesses the computational efforts from niche groups of experts rather than the `faceless crowd' \cite{de2012nichesourcing}. Nichesourcing combines the strengths of the crowd with those of professionals \cite{de2012nichesourcing,oosterman2014crowd}. In our case we organized several data collection sessions with NGO operators, who are trained experts, specialized in writing counter-narratives that are meant to fight hatred and de-escalate the conversation. In this way we build the first large-scale, multilingual, publicly available, expert-based dataset of hate speech/counter-narrative pairs for English, French and Italian, focusing on the hate phenomenon of \textit{Islamophobia}.  The construction of this dataset involved more than $100$ operators and more than $500$ person-hours of data collection. After the data collection phase, we hired three non-expert annotators, that performed additional tasks that did not require specific domain expertise ($200$ person-hours of work): paraphrase original hate content to augment the number of pairs per language, annotate hate content sub-topic and counter-narrative type, translate content from Italian and French to English to have parallel data across languages. This additional annotation grants that the dataset can be used for several NLP tasks related to hate speech. 

The remainder of the paper is structured as follows. First, we briefly %discuss the concept of counter-narratives (Section \ref{counter_narr}), then in Section \ref{related_work}, we discuss the 
discuss related work on hate speech in Section \ref{related_work}. Then, in Section \ref{dataset}, we introduce our CONAN dataset and some descriptive statistics, followed by a quantitative and qualitative analysis on our dataset in Section \ref{evaluation}. We conclude with our future works in Section \ref{future_work}.

\section{Related Work} \label{related_work}

With regard to hatred online, we will focus on three research aspects about the phenomenon, i.e. (i) publicly available datasets, (ii) methodologies for detecting hate speech, (iii) seminal works that focus on countering hate speech. 

\paragraph{Hate datasets.} Several hate speech datasets are publicly available, usually including a binary annotation, i.e. whether the content is hateful or not \cite{reynolds2011using, rafiq2015careful, hosseinmardi2015detection, de2018hate, elsherief2018peer}. Also, several shared tasks have released their datasets for hate speech detection in different languages. For instance, there is the German abusive language identification on SMPs at Germeval \cite{bai2018rug}, or the hate speech and misogyny identification for Italian at EVALITA \cite{del2017hate, fersini2018overview} and for Spanish at IberEval \cite{ahluwalia2018detecting, shushkevich2018classifying}. Bilingual hate speech datasets are also available for Spanish and English \cite{pamungkas201814}. %However, certain crucial information is not available, such as hate speech/counter-narrative pairs and various categories of hate speech as well as multilinguality. 

% Hatebase\footnote{https://www.hatebase.org/} is a usage-based repository created via crowdsourcing, offering multilingual and location-based vocabulary for hate speech. These features allow for several tasks, such as hate speech classification \cite{nobata2016abusive, davidson2017automated}, monitoring migration and hatred in different cultures and regions\footnote{https://thesentinelproject.org/2013/03/25/introducing-hatebase-the-worlds-largest-online-database-of-hate speech/}. Nevertheless, Hatebase is keyword-based, thus being inadequate to differentiate hate speech from counter-narrative if both contain hate keywords \cite{saleem2017web,wiegand2018inducing}.

%\footnote{39\% of tweets was lost when we downloaded the dataset from \citet{mathew2018analyzing} in December, 2018}.  %\sst{Can we put a footnote to at least one of the following examples and write down when we download their data and what percentage of it was already lost on that date? \textbf{YES!}} 
%\mg{EITHER CHECK LOSS FOR THE TWO FOLLOWING CITATION OR REMOVE "for istance"}
\citet{waseem2016hateful} released 16k annotated tweets containing 3 offense types: sexist, racist and neither. % In one year after the dataset was released, 4.16\% of hate content was already lost \cite{Klubicka2018}.
\citet{ross2017measuring} first released a German hate speech dataset of 541 tweets targeting refugee crisis and then offered insights for the improvement on hate speech detection by providing multiple labels for each hate speech. 

It should be noted that, due to the copyright limitations, usually hate speech datasets are distributed as a list of tweet IDs making them ephemeral and prone to data loss \cite{Klubicka2018}. For this reason, \citet{sprugnoli2018creating} created a multi-turn annotated WhatsApp dataset for Italian on Cyberbullying, using simulation session with teenagers to overcome the data collection/loss problem.  

%but also addressed the importance of building a living lab to study cyberbullying interaction among teenagers.

\paragraph{Hate detection.} Several works have investigated online English hate speech detection and the types of hate speech. Owing to the availability of current datasets, researchers often use supervised-approaches to tackle %Several works have been using machine learning algorithms on
hate speech detection on SMPs including blogs \cite{warner2012detecting, djuric2015hate, gitari2015lexicon}, Twitter \cite{xiang2012detecting, silva2016analyzing, mathew2018analyzing}, Facebook \cite{del2017hate}, and Instagram \cite{zhong2016content}. The predominant approaches are to build a classifier trained on various features derived from lexical resources \cite{gitari2015lexicon, burnap2015cyber, burnap2016us}, n-grams \cite{sood2012automatic, nobata2016abusive} and knowledge base \cite{dinakar2012common}, or to utilize deep neural networks \cite{mehdad2016characters, badjatiya2017deep}. In addition, other approaches have been proposed to detect subcategories of hate speech such as anti-black \cite{kwok2013locate} and racist \cite{badjatiya2017deep}. \citet{silva2016analyzing} studied the prevalent hate categories and targets on Twitter and Whisper, but limited hate speech only to the form of \textit{I  $<$intensity$>$  $<$user intent$>$ $<$any word$>$}. A comprehensive overview of recent approaches on hate speech detection using NLP can be found in \cite{schmidt2017survey,fortuna2018survey}.

%\mg{Yiling: more works on hate detection. SemEval winning systems? The review I forwarded to you?}

\paragraph{Hate countering.} Lastly, we should mention that a very limited number of studies have been conducted on counter-narratives \cite{benesch2014countering, schieb2016governing, ernst2017hate, mathew2018thou}.
\citet{mathew2018thou} collected Youtube comments that contain counter-narratives to YouTube videos of hatred. \citet{schieb2016governing} studied the effectiveness of counter-narrative on Facebook via a simulation model. The study of \citet{wright2017vectors} shows that some arguments among strangers induce favorable changes in discourse and attitudes. To our knowledge, there exists only one very recent seminal work \cite{mathew2018analyzing}, focusing on the idea of collecting hate message/counter-narrative pairs from Twitter. They used a simple pattern in the form \textit{(I $<$hate$>$ $<$category$>$)} to first extract hate tweets and then manually annotate counter-narratives found in the responses. Still, there are several shortcomings of their approach: (i) this dataset already lost more that 60\% of the pairs in a small time interval (content deletion) since only tweet IDs are distributed, (ii) it is only in English language, (iii) the dataset was collected from a specific template which limits the coverage of hate speech, and (iv) many of these answers come from ordinary web users and contain -for example- offensive text, that do not meet the de-escalation intent of NGOs and the standards/quality of their operators' responses.  \\

Considering the aforementioned works, we can reasonably state that no suitable corpora of counter-narratives is available for our purposes, especially because the natural `countering' data that can be found on SMP -- such as example \ref{ex_c} -- often does not meet the required standards. For this reason we decided to build CONAN, a dataset of COunter NArratives through Nichesourcing. 

%More precisely, collecting real counter-narrative examples to train end to end systems is a non trivial task, especially because the natural `countering' data that can be found on SMP -- such as example \ref{ex_c} -- often does not meet the required standards.

%\mg{Tonelli's work on whatsApp dataset for Bulling: attacker defender bullied, all simulated in Italian}

\section{CONAN Dataset} \label{dataset}

In this section, we describe the characteristics that we intend our dataset to posses, the nichesourcing methodology we employed to collect the data and the further expansion of the dataset together with the annotation procedures. Moreover, we give some descriptive statistics and analysis for the collected data. CONAN can be downloaded at the following link \url{https://github.com/marcoguerini/CONAN}.

\subsection{Fundamentals of the Dataset}

Considering the shortcomings of the existing datasets and our aim to provide a reliable resource to the research community, we want CONAN to comply with the following characteristics: 

\paragraph{Copy-free data.} We want to provide a dataset that is not ephemeral, by releasing only copy-free textual data that can be directly exploited by researches without data loss across time, as originally pointed out in \cite{Klubicka2018}.

\paragraph{Multilingual data.} Our dataset is produced as a multilingual resource to allow for cross lingual studies and approaches. In particular, it contains hate speech/counter-narrative pairs for English, French, and Italian.

\paragraph{Expert-based data.}
%The hat data distributed has been provided by three different NGOs. Responses are expert based, composed by operators from non-governmental organizations to each hate speech. 
The hate speech/counter-narrative pairs have been collected through nichesourcing to three different NGOs from United Kingdom, France and Italy. Therefore, both the responses and the hate speech itself are expert-based and composed by operators, specifically trained to oppose online hate speech.

\paragraph{Protecting operator's identity.} 
We aim to create a secure dataset that will not disclose the identity of operators in order to protect them against being tracked and attacked online by hate spreaders. This might be the case if we were to collect their real SMP activities, following a procedure similar to the one in \citet{mathew2018analyzing}. %Disclosing operators and free-speech ambient where everyone can speak their minds anonymously. 
Therefore our data collection was based on simulated SMP activity. %, protecting operator's identity by asking only demographic profile but not personal information (see next paragraph).

\paragraph{Demographic-based metadata.}
Demographic-based NLP can be used for several tasks, such as characterizing personal linguistic styles \cite{johannsen2015cross, hovy2016social, van2018bleaching, dell2018overview}, improving text classification \cite{mandel2012demographic, volkova2013exploring, hovy2015demographic}, or personalizing conversational agents \cite{qiu2010study, mazare2018training}. In this work, we collect demographic information of participants; i.e. gender, age, and education level, to provide data for counter-narrative personalization. 
%\citet{hovy2015demographic} improved text classification task by including demographic information like gender and age, compared to demographic-agnostic model. 
%\citet{johannsen2015cross} investigated syntactic variation among demographic groups and found age- and gender-specific syntactic variations across languages. 
%Following the previous work, we would like to address demographic adaptation 
%\cite{hovy2015demographic} in hate speech and personalized counter-narrative generation\sst{Are we really addressing it or just as a future work, suggesting it??}, although few studies yet examine them deeply. 
%Accurate author profiling and identification of sentiment polarity can help characterize hate users and provide personalized counter-narrative. In order to elicit more potential applications from contextual information, in this work we collect demographic data from users. \citeauthor{van2018bleaching} \citeyear{van2018bleaching} used bleached model with abstract features for cross-lingual gender prediction and achieve better performance than lexical models. 

%\paragraph{WHERE???? more than a DB we want an 'ecosystem'} 

\subsection{Dataset Collection}
We have followed the same data collection procedure for each language to grant the same conditions and comparability of the results. The data collection has been conducted along the following steps:\\

\noindent\textit{1. Hate speech collection.} For each language we asked two native speaker experts (NGO trainers) to write around 50 prototypical islamophobic short hate texts. This step was used to ensure that: (i) the sample uniformly covers the typical `arguments' against Islam as much as possible, (ii) we can distribute to the NLP community the original hate speech as well as its counter-narrative. % text to which the counter-narrative is a response. %In particular, for Italian, we further collected hate text for 3 other hate target communities: Gypsies, LGBTI, and Migrants and refugees.
%overcoming the copyright issues.

\noindent\textit{2. Preparation of data collection forms.} We prepared three online forms (one per language) with the same instructions for the operators translated in the corresponding language. For each language, we prepared 2 types of forms: in the first users can respond to hate text prepared by NGO trainers, in the second users can write their own hate text and counter-narratives at the same time. In each form  operators were first asked to anonymously provide their demographic profile including age, gender, and education level; secondly to compose up to 5 counter-narratives for each hate text.

%provided either by NGO trainers or by themselves. 

\noindent\textit{3. Counter-narrative instructions.} \label{CN_instruction} The operators were already trained to follow the guidelines of the NGOs for creating proper counter-narratives. Such guidelines are highly consistent across languages and across NGOs, and are similar to those in `Get the Trolls Out' project\footnote{http://stoppinghate.getthetrollsout.org/}. These guidelines emphasize using fact-bounded information and non-offensive language in order to avoid escalating the discussion as outlined in Table \ref{table:CN_guidelines}. Furthermore, for our specific data collection task, operators were asked to follow their intuitions without over-thinking and to compose reasonable responses. The motivation for this instruction was to collect as much and as diverse data as possible, since for current AI technologies (such as deep learning approaches) quantity and quality are of paramount importance and few perfect examples do not provide enough generalization evidence. Other than this instruction and the fact of using a form -- instead of responding on a SMP -- operators carried out their normal counter messaging activities.

%\paragraph{(1)}Don't be abusive\\
%Before submitting a response, make sure the response does not spread any hate, bigotry, prejudice or illegal content. We want to maintain the conversations peaceful and not to degenerate into a conflict. We are talking about people not categories. 

%\paragraph{(2)}Don't spread hate\\

%\paragraph{(2)}Think about the objectives.\\ 
%Before writing a response, think about the effect it may create and the one you want to obtain. Paying attention to the objectives will help compose responses using proper words.

%\paragraph{(3)}Call for influential users\\
%Enlisting influential supporters (civic leaders, politicians, subject experts) will help bring attention and deepen the effect to counter-narrative. %Hate speech can be intervened by including a fair comparison between citizens.

%\paragraph{(4)}Use credible evidence\\
%The information in hate speech may be confusing and misleading. We ask for clarification when that is the case. Counter it with credible evidence and knowledge from reliable sources.

%\paragraph{(5)}Think about the tone\\
%We can demonstrate understanding and support to those who might be attacked. Be careful of using sarcasm, humour, parody and satire. We can use them, if we are able to master it as they run the danger of being antagonistic.

\begin{table*}[t!]
\centering

\begin{tabularx}{\textwidth}{|X|}
  %Guideline & Description \\
  \hline
  \textbf{Don't be abusive} Before submitting a response, make sure the response does not spread any hate, bigotry, prejudice or illegal content. We want to maintain the conversations peaceful and not to degenerate into a conflict. We are talking about people not categories.\\

  %Don't spread hate & \\
  %\hline
  \textbf{Think about the objectives} Before writing a response, think about the effect it may create and the one you want to obtain. Paying attention to the objectives will help %compose responses 
  use proper words. \\

 \textbf{Call for influential users} Enlisting influential supporters (civic leaders, politicians, subject experts) will help bring attention and deepen the effect to counter-narrative.\\

  \textbf{Use credible evidence} The information in hate speech may be confusing and misleading. Ask for clarification when %that is the case
  necessary. Counter it with credible evidence and %knowledge from 
  use reliable sources. \\

  \textbf{Think about the tone } We can demonstrate understanding and support to those who might be attacked. Be careful of using sarcasm, humour, parody and satire. We can use them, if we are able to master it as they run the danger of being antagonistic.\\
     \hline
\end{tabularx}
\caption{Counter-narrative guidelines adopted by the three NGOs.}
  \label{table:CN_guidelines}

\end{table*}

\noindent\textit{4. Data collection sessions.} For each language, we performed three data collection sessions on different days. Each session lasted roughly three hours\footnote{Duration was decided during a pilot session: after three hours operators were no longer effective in producing counter-narratives, since it is a cognitively fatiguing task.} and had a variable number of operators -- usually around 20 (depending on their availability). Operators are different from NGO trainers and might change across sessions.   
Operators were gathered in the same room (NGO premises) with a computer, and received a brief introduction from the NGO trainer. This introduction was about our specific counter-narrative collection task, as described above. %Later, operators were instructed to compose counter-narratives using the online forms and to browse for relevant material if needed, such as statistical reports and factual information. 
A sample of the collected data for the three languages is given in Table \ref{table:pairs_examples}.

\begin{table*}[t!]
\centering
%\begin{tabular}{lll}
\begin{tabular}{p{.33\textwidth}p{.61\textwidth}}
Hate Speech & Counter-Narrative \\
\hline
Every Muslim is a potential terrorist. & Every Muslim is also a potential peacemaker, doctor, philanthropist... What's your point? \\
\hline
%Islam is just a repressive ideology that persecutes women and gay people. & Women and gay people have been persecuted by many groups all over the world and it needs to change. Is there a particular reason you are picking on Islam?\\
%\hline
%Gli islamici sono tutti terroristi. & La colpa di pochi non pu\`{o} ricadere su tutti gli islamici, inoltre sono proprio loro ad essere le prime vittime del terrorismo.\\
%\hline
I musulmani violentano le nostre donne e vanno castrati.  & Buonasera, se ho capito bene, lei mi sta dicendo che tutti gli uomini adulti di fede islamica presenti in Italia, hanno violentato, violentano o violenteranno le donne italiane? Pu\`{o} portare dei dati a supporto della sua tesi? \\
\textit{Muslims rape our women and must be neutered.} & \textit{Good evening, if I understood correctly, you're telling me that all adult men of Islamic faith present in Italy, raped, rape or are going to rape Italian women? Can you bring in data in support of your statement?} \\
\hline
%Les musulmans sont tous des terroristes & Islam n'est pas une religion qui pr\^{o}ne ou d\'{e}fend le terrorisme.\\
%\hline
Le voile est contraire \`{a} la la\"{\i}cit\'{e}. & Bien au contraire la la\"{\i}cit\'{e} permet \`{a} tout citoyen de vivre librement sa confession.\\
\textit{The veil is contrary to secularism.} & \textit{On the contrary, secularism allows every citizen to freely profess his faith.}\\
\hline
\end{tabular}
\caption{Example pairs for the three languages, along with English translations.}
  \label{table:pairs_examples}
\end{table*}

%\sst{Let's all be careful about "plural nouns in a row". The most grammatically correct one (except the reaaally specific situations)  is using only the stressed noun as plural: "pair examples" instead of "pairs examples"}
%\paragraph{Data collection `numbers'.} In total we had more than \textbf{400} hours of data collection, with roughly \textbf{80} annotators over the three languages. We collected more than \textbf{3000} pairs of hate / counter-narratives. Each counter-narrative needed \textbf{7/8} minutes on average to be composed.   

%Going into details of hate speech/counter-narrative pairs:  we collected 1092 pairs for Italian, 1290 pairs for English, and \textbf{XX} pairs for French, for a total of \textbf{XX} pairs written by \textbf{14+36+xx} operators. 

%\mg{demographic profile??}

% Specifically, for Italian dataset, all the operators worked and discussed together to compose consensus counter-narratives, apart from counter-narratives made by individuals. We labeled the consensus counter-narratives as individual responses created by a consensus author.

%\subsection{demographic profile}

\subsection{Dataset Augmentation and Annotation}

After the data collection phase, we hired three non-expert annotators, that performed additional work that did not require specific domain expertise. Their work amounted to roughly 200 hours. In particular they were asked to (i) paraphrase original hate content to augment the number of pairs per language, (ii) annotate hate speech sub-topics and counter-narrative types (iii) translate content from French and Italian to English to have parallel data across languages. To guarantee data quality, after the annotation and the augmentation phase, a validation procedure has been conducted by NGO trainers on the newly generated data for their specific language.

\paragraph{Paraphrasing for augmenting data pairs.} 
Recent deep learning approaches are data hungry, and data augmentation is a way to mitigate the problem. For instance, to improve text classification performance for sexism, new tweets are generated by replacing words in original tweets with synonyms from ConceptNet \cite{sharifirad2018boosting}. Other examples of data augmentation strategies are back translation \cite{P16-1009} and gold standard repetition \cite{chatterjee2017guiding} that have been used in sequence-to-sequence Machine Translation. 
%In the study by \citet{ding2018generating}, data inversion was used to artificially generate keywords given a large quantity of questions. 
In all these tasks, adding the synthetic pairs to the original data always results in significant improvements in the performance.

In line with the idea of artificially augmenting pairs, and since in our dataset we have many responses for few hate speeches, we %managed to effectively 
produced two manual paraphrases of each hate speech and paired them with the counter-narratives of the original one. Therefore
we increased the number of our pairs by three times in each language.

\paragraph{Counter-narrative type annotation.} In this task, we asked the annotators to label each counter-narrative with types. %When deciding on the counter-narrative type, the annotators did not considered it in isolation, but together with the original hate statement so to have a better context for classification.  
Based on the counter-narrative classes proposed by \cite{benesch2016,mathew2018thou}, we defined the following set of types: \textsc{Presentation of Facts}, \textsc{Pointing out hypocrisy or contradiction}, \textsc{Warning of consequences}, \textsc{Affiliation}, \textsc{Positive Tone}, \textsc{Negative Tone}, \textsc{Humor}, \textsc{Counter-Questions}, \textsc{Other}. With respect to the original guidelines, we added a new type of counter-narrative called \textsc{Counter-Questions} to cover expressions/replies using a question that can be thought-provoking or asking for more evidence from the hate speaker. In fact, a preliminary analysis showed that this category is quite frequent among operator responses. Finally, each counter-narrative can be labeled with more than one type, thus making the annotation more fine-grained. 

Two annotators per language annotated all the counter-narratives independently.
A reconciliation phase was then performed for the disagreement cases. 

\paragraph{Hate speech sub-topic annotation.} We labeled sub-topics of hate content to have an annotation that can be used both for fine grained hate speech classification, and for exploring the correlation between hate sub-topics and counter-narrative types. %In addition, we annotated the polarity of hate content, which can be applied to sentiment analysis to quantify hate emotion, identify relevant topics \cite{xu2012learning}, and generate appropriate counter-narratives. \mg{what are these categories, where do they come from?} 
The following sub-topics are determined for the annotation based on the guidelines used by NGOs to identify hate messages (mostly consistent across languages):
\textsc{Culture}, criticizing Islamic culture or particular aspects such as religious events or clothes; 
\textsc{Economics}, hate statements about Muslims taking European workplaces or not contributing economically to the society; 
\textsc{Crimes}, hate statements about Muslims committing actions against the law;
\textsc{Rapism}, a very frequent topic in hate speech, for this reason it has been isolated from the previous category;
\textsc{Terrorism}, accusing Muslims of being terrorists, killers, preparing attacks;
\textsc{Women Oppression}, criticizing Muslims for their behavior against women; 
\textsc{History}, stating that we should hate Muslims because of historical events;
\textsc{other/generic}, everything that does not fall into the above categories.

As before, two annotators per language annotated all the material. Also in this annotation task, a reconciliation phase was performed for the disagreement cases. 

\paragraph{Parallel corpus of language pairs.}
To allow studying cross-language approaches to counter-narratives and more generally to increase language portability, we also translated the French and the Italian pairs (i.e. hate speech and counter-narratives) to English. Similar motivations can be found in using zero-short learning to translate between unseen language pairs during training \cite{johnson2017google}. %, or to recognize activities in videos for out-of-domain actions\sst{highly-irrelevant} \cite{guadarrama2013youtube2text}. 
With parallel corpora we can exploit cross-lingual word embeddings %within the counter-narrative generation system, which 
to enable knowledge transfer between languages \cite{schuster2018cross}. %\sst{cite about knowledge transfer between languages?} %Furthermore, it is promising to learn multilingual sentence representations using a single model \cite{artetxe2018massively} and apply it to counter-narrative generation in dialog system.

\subsection{Dataset Statistics}

In total we had more than 500 hours of data collection with NGOs, where we collected 4078 hate speech/counter-narrative pairs; specifically, 1288 pairs for English, 1719 pairs for French, and 1071 pairs for Italian. At least 111 operators participated in the 9 data collection sessions and each counter-narrative needed about 8 minutes on average to be composed. The paraphrasing of hate messages and the translation of  French and Italian pairs to English brought the total number of pairs to more than 15 thousand. Regarding the token length of counter-narratives, we observe that there is a consistency across the three languages with 14 tokens on average for French, and 21 for Italian and English. Considering counter-narrative length in terms of characters, only a small portion (2\% for English, 1\% for French, and 5\% for Italian) contains more than 280 characters, which is the character limit per message in Twitter, one of the key SMPs for hate speech research. Further details on the dataset can be found in Table \ref{table:DS_statistics}. 

% We collected more than 4000 pairs of hate / counter-narratives. Each counter-narrative needed \textbf{7/8} minutes on average to be composed.   Going into details of hate speech/counter-narrative pairs: We collected 1074 pairs for Italian, 1288 pairs for English, and \textbf{1719} pairs for French, for a total of \textbf{4081} pairs written by at least 111 \yl{14+35+62} operators. The paraphrasing of hate speeches and the translation of the French and English pairs to English  brings the total number \textbf{above ten thousand} pairs. With regard to counter-narratives length we can see that it is quite consistent across languages, ranging from 14 tokens on average for French to 21 for Italian. Considering counter-narrative length in terms of chars -- that is relevant for SMPs such as Twitter, that give a cap on their total number per message -- we see that only a small portion (2\% for English, 1\% for French, and 5\% for Italian) is above the threshold of 280 chars. Further details on the collected data set can be found in Table \ref{table:DS_statistics}. 

\begin{table}[t!]
\centering
\begin{tabular}{p{2.7cm}rrr}%{lrrr}p{7.5cm}
  & English & French & Italian\\
  \hline
  original pairs & 1288 & 1719 & 1071 \\
  augmen. pairs & 2576 & 3438 & 2142 \\
  transl. pairs & 2790 & - & - \\
  total pairs & 6654 & 5157 & 3213 \\
  \hline
  HS & 136 & 50 & 62\\
  CN\_per\_HS$_\mu$ & 9.47 & 34.38 & 17.27\\
  CN\_per\_HS$_{sd}$ & 7.56 & 53.86 & 26.48\\
  \hline
  HS vocabulary & 947 & 193 &  343\\
  HS$+$aug. vocab. & 1631 & 333 & 790\\
  CN vocabulary & 3556 & 4018 & 3728\\
  \hline
  HS words & 2950 & 434 &  751\\
  HS$+$aug. words & 9770 & 1172 & 2633\\
  CN words & 27677 & 23730 & 23129\\
  \hline
  HS\_words$_\mu$ & 21.69 & 8.68 & 12.11\\
  HS\_words$_{sd}$ & 10.29 & 4.02 & 6.69\\
  
  HS$+$aug.\_words$_\mu$ & 18.72 & 5.31 & 14.16\\
  HS$+$aug.\_words$_{sd}$ & 10.05 & 4.73 & 7.65\\
  CN\_words$_\mu$ & 21.49 & 13.80 & 21.60\\
  CN\_words$_{sd}$ & 11.06 & 11.44 & 12.42\\
  %responses $<$ 140 characters & & & \\
  %responses $>$ 140 characters & 464 & 50 & 410\\
  %responses $>$ 280 characters & 27 & 5 & 57\\
   
   %words$_\mu$ &&&\\
   %words$_{sd}$ &&&\\
     \hline
\end{tabular}
\caption{Main statistics of the dataset. HS stands for Hate Speech, CN stands for Counter-Narrative.}
  \label{table:DS_statistics}
\end{table}
Regarding demographics, the majority of responses were written by operators that held a bachelor's or a higher degree (95\% for English, 65\% for French, and 69\% for Italian). As it is shown in Table \ref{table:Dem_statistics}, there is a good balance in responses with regard to declared gender, with a slight predominance of counter-narratives written by female operators in English and Italian (53 and 55 per cent respectively) while a slight predominance of counter-narratives written by male operators is present in French (61\%). %57\%
Finally, the predominant age bin is 21-30 for English and Italian, while for French is in the range 31-40. 

\begin{table}[h!]
\centering
\begin{tabular}{lrrr}
  & EN & FR & IT\\
  \hline
  \  $<$ high school & - & 5\% & 14\%\\
  \ high school & - & 14\% & 10\%\\
  \ $<$ university & 5\% & 16\% & 6\%\\
  \ bachelor & 51\% & 17\% & 34\%\\
  \ master & 44\% & 35\% & 30\%\\
  \ PhD & - & 13\% & 5\%\\
     \hline
   female & 53\% & 39\% & 55\%\\
   male & 47\% & 61\% & 45\%\\
     \hline
$< =$ 20 & - & - & 15\%\\
21 - 30  & 74\% & 15\% & 42\%\\
31 - 40  & - & 51\% & 7\%\\
41 - 50  & 18\% & 20\% & 15\%\\
51 - 60  & - & 11\% & 16\%\\ 
$>$ 60   & 8\% & 3\% & 5\%\\ 
\hline
%missing value \mg{how is it possible for English?}   &&& \\ \yiling{That came from Lisa's responses, and I didn't know her age before. Now I know. I comment out this line}
\end{tabular}
\caption{Demographic profile of the operators.}
  \label{table:Dem_statistics}
\end{table}
\begin{table}[h!]
\centering
\begin{tabular}{lrrr}
Type & EN & FR & IT \\
\hline
affiliation  &  1 & 4 & 1 \\
consequences &  0 & 1 & 0 \\
denouncing  &  19 & 18 & 13 \\
facts     &  38 & 37 & 47 \\
humor     &  8 &  6 & 5 \\
hypocrisy   &  16 & 14 & 10 \\
negative   &  0 & 0 & 0 \\
other   &  0 & 4 & 1 \\
positive   &  6 & 5 & 7 \\
question   &  12 & 11 & 16 \\
\hline
\end{tabular}
\caption{Counter-narrative type distribution over the three languages (\% over the total number of labels).} %\yl{Should we display the languages in the order of EN-FR-IT, as the rest of the tables?}
  \label{table:CN_types_distribution}
\end{table}

Considering the annotation tasks, we give the distribution of counter-narrative types per language in Table \ref{table:CN_types_distribution}. As can be seen in the table, there is a consistency across the languages such that \textsc{Facts}, \textsc{Question}, \textsc{Denouncing}, and \textsc{hypocrisy} are the most frequent counter-narrative types. 
%Considering the counter-narrative typology annotation,
Before the reconciliation phase, the agreement between the annotators was moderate: Cohen's Kappa\footnote{Computed using Mezzich's methodology to account for possible multiple labels that can be assigned to a text by each annotator \cite{mezzich1981assessment}.} 0.55 over the three languages. This can be partially explained %also
by the complexity of the messages, that often fall under more than one category (two labels were assigned in more than 50\% of the cases). 
On the other hand, for hate speech sub-topic annotation, the agreement between the annotators was very high even before the reconciliation phase (Cohen's Kappa 0.92 over the three languages). A possible reason is that such messages represent short and prototypical hate arguments, as explicitly requested to the NGO trainers. In fact, the vast majority has only one label. In Table~\ref{table:HS_types_distribution}, we give a distribution of hate speech sub-topics per language. As can be observed in the table, the labels are distributed quite evenly among sub-topics and across languages - in particular, \textsc{Culture}, \textsc{Islamization}, \textsc{Generic}, and \textsc{Terrorism} are the most frequent sub-topics.

\begin{table}[h!]
\centering
\begin{tabular}{lrrr}
Type & EN & FR & IT \\
\hline
crimes & 10 & 0 & 7 \\
culture & 30 & 26 & 11 \\
economics & 4 & 1 & 8 \\
generic & 20 & 27 & 8 \\
islamization & 11 & 7 & 36 \\
rapism & 15 & 0 & 7 \\
terrorism & 6 & 14 & 19 \\
women & 4 & 25 & 4 \\
\hline
\end{tabular}
\caption{hate speech sub-topic type distribution over the three languages (\% over the total number of labels).}
  \label{table:HS_types_distribution}
\end{table}

\section{Evaluation} \label{evaluation}

In order to assess the quality of our dataset, we ran a series of preliminary experiments that involved three annotators to judge hate speech/counter-narrative pairs along a yes/no dimension. % These experiments were run with the possible development of automatic suggestion tools for NGO's operators in mind. 

\paragraph{Augmentation reliability.} The first experiment was meant to assess how natural a pair is when coupling a counter-narrative with the manual paraphrase of the original hate speech it refers to. We administered 120 pairs to the subjects to be evaluated: 20 were kept as they are so to have an upper bound representing \textsc{Original} pairs. In 50 pairs we replaced the hate speech with a \textsc{Paraphrase}, and in the 50 remaining pairs, we randomly matched a hate speech with a counter-narrative from another hate speech (\textsc{Unrelated} baseline). Results show that 85\% of the times in the \textsc{Original} condition hate speech and counter-narrative were considered as clearly tied, followed by the 74\% of times by  \textsc{Paraphrase} condition, and only 4\% of the \textsc{Unrelated} baseline, this difference is statistically significant with $p<.001$ (w.r.t. $\chi^2$ test). This indicates that the quality of augmented pairs is almost as good as the one of original pairs.

\paragraph{Augmentation for counter-narrative selection.} Once we assessed the quality of augmented pairs, we focused on the possible contribution of the paraphrases also in standard information retrieval approaches that have been used as baselines in dialogue systems \cite{lowe2015ubuntu,Mazar2018TrainingMO}. We first collected a small sample of natural/real hate speech from Twitter using relevant keywords (such as ``stop Islam") and manually selected those that were effectively hate speeches. We then compared 2 tf-idf response retrieval models by calculating the tf-idf matrix using the following document variants: (i) hate speech and counter-narrative response, (ii) hate speech, its 2 paraphrases, and counter-narrative response. The final response for a given sample tweet is calculated by finding the highest score among the cosine similarities between the tf-idf vectors of the sample and all the documents in a model.

For each of the 100 natural hate tweets, we then provided 2 answers (one per approach) selected from our English database. Annotators were then asked to evaluate the responses with respect to their relevancy/relatedness to the given tweet. %(score from 0 to 2 , "0" not at all, "2" very relevant). 
Results show that introducing the augmented data as a part of the tf-idf model provides $9\%$ absolute increase in the percentage of the agreed `very relevant' responses, i.e. from $18\%$ to $27\%$ - this difference is statistically significant with $p<.01$ (w.r.t. $\chi^2$ test). This result is especially encouraging since it shows that the augmented data can be helpful in improving even a basic automatic counter-narrative selection model.

\paragraph{Impact of Demographics.} The final experiment was designed to assess whether demographic information can have a beneficial effect on the task of counter-narrative selection/production. In this experiment, we selected a subsample of 230 pairs from our dataset written by 4 male and 4 female operators that were controlled for age (i.e. same age range). We then presented our subjects with each pair in isolation and asked them to state whether they would definitely use that particular counter-narrative for that hate speech or not. Note that, in this case, we did not ask whether the counter-narrative was relevant, but if they would use that given counter-narrative text to answer the paired hate speech. The results show that in the \textsc{SameGender} configuration (gender declared by the operator who wrote the message and gender declared by the annotator are the same), the appreciation was expressed 47\% of the times, while it decreases to 32\% in the \textsc{DifferentGender} configuration (gender declared by the operator who wrote the message and gender declared by the annotator are different). This difference is statistically significant with $p<.001$ (w.r.t. $\chi^2$ test), indicating that even if operators were following the same guidelines and were instructed on the same possible arguments to build counter-narratives, there is still an effect of their gender on the produced text, and this effect contributes to the counter-narrative preference in a \textsc{SameGender} configuration.

%Additional experiments: HUMAN EVALUATION of synthetic pairs  !!!
%HUMAN EVALUATION of translated pairs ???

%\section{Data Ecosystem} \label{ecosystem}

%This data useful for many NLP approaches/tasks:
%e.g. porting from another language (parallel data)
%e.g. porting to other hate-targets (not only islam target)
%e.g. personalized CN generation
%...

\section{Conclusion} \label{future_work}
% and Future Work
As online hate content rises massively, responding to it with counter-narratives as a combating strategy draws the attention of international organizations. Although a fast and effective responding mechanism can benefit from an automatic %counter-narrative 
generation system, the lack of large datasets of appropriate counter-narratives hinders tackling the problem through supervised approaches such as deep learning. 
In this paper, we described CONAN: the first large-scale, multilingual, and expert-based hate speech/counter-narrative dataset for %in 
English, French, and Italian. The dataset consists %of more than 1k hate speech/counter-narrative pairs for each language, in total
of 4078 pairs over the 3 languages. Together with the collected data we also provided several types of metadata: expert demographics, hate speech sub-topic and counter-narrative type. Finally, we expanded the dataset through translation and paraphrasing. %Finally, we provide initial experiments to evaluate the quality of our data. 

%The dataset provides several insights on further research. 
As future work, we intend to continue collecting more data for Islam and to include %several hate speech topics to the dataset 
other hate targets such as migrants or LGBT$+$, in order to put the dataset at the service of other organizations and further research. % Moreover, we plan to add argumentative label\sst{What does this mean? Argumentative label seems like the argument type that we have already annotated for the counter narratives.} to each hate content in order to investigate the..  how this improve the classification.\sst{What classification??} 
Moreover, as a future direction, we want to utilize CONAN dataset to  
% effective strategies for composing counter-narratives 
develop a counter-narrative generation tool that can support NGOs in fighting hate speech online, considering counter-narrative type as an input feature.

%utilizing the dataset that we have described throughout this paper will allow us to develop a conversational agent that can pursue a dialogue by generating counter-narratives when it encounters a hate speech, hence supporting the organizations that fight hate speech online as an automatic aid
%others for different communities, for instance, migrants, LGBT, and gypsies. For this we can further analyze the influence of various strategies of counter-narrative on various hate targets. Another direction could be to add argumentative label to each hate content and investigate how this improve the classification.\sst{What classification??} Lastly, one interesting objective is to develop a conversational agent to generate counter-narratives when it detects hate speech and hence monitoring and correcting hateful content on social media.

%- keep on gathering data
%- adding argumentative label
%- real generation 

\section*{Acknowledgments}
This work was partly supported by the HATEMETER project within the EU Rights, Equality and Citizenship Programme 2014-2020.  
We are grateful to the following NGOs and all annotators for their help: Stop Hate UK, Collectif Contre l'Islamophobie en France, Amnesty International (Italian Section - Task force  hate speech).

\bibliography{acl2019}
\bibliographystyle{acl_natbib}

%\appendix

%\section{Appendices}
%\label{sec:appendix}

%\section{Supplemental Material}
%\label{sec:supplemental}

\end{document}